# FST Based Morphological Analyzer for Hindi Language

Deepak Kumar[1], Manjeet Singh[2], and Seema Shukla[3]

[1]Department of Information Technology, JSS Academy of Technical Education Noida, Uttar Pradesh, India
*deepakk799@gmail.com*

[2]Department of Information Technology, JSS Academy of Technical Education Noida, Uttar Pradesh, India
*manjeet207@gmail.com*

[3]Department of Computer Science & Engineering, JSS Academy of Technical Education Noida, Uttar Pradesh, India
*seemashukla@gmail.com*

## Abstract

Hindi being a highly inflectional language, FST (Finite State Transducer) based approach is most efficient for developing a morphological analyzer for this language. The work presented in this paper uses the SFST (Stuttgart Finite State Transducer) tool for generating the FST. A lexicon of root words is created. Rules are then added for generating inflectional and derivational words from these root words. The Morph Analyzer developed was used in a Part Of Speech (POS) Tagger based on Stanford POS Tagger. The system was first trained using a manually tagged corpus and MAXENT (Maximum Entropy) approach of Stanford POS tagger was then used for tagging input sentences. The morphological analyzer gives approximately 97% correct results. POS tagger gives an accuracy of approximately 87% for the sentences that have the words known to the trained model file, and 80% accuracy for the sentences that have the words unknown to the trained model file.

## 1. Introduction

In recent years, large collections of digitized Hindi documents were created due to the improvement in Information Technology. An efficient accessing system has to take into account the morphology of the language through a systematic linguistic study in order to reveal words that are significant to users, such as historians, linguists etc and describe morpho-phonological rules. The result of that study is the starting point for the construction of computational morphologies providing ability to search documents using word root and locate all the corresponding inflected words[1,2].

A highly inflectional language has the capability of generating hundreds of words from a single root. Hence, morphological analysis is vital for high level applications to understand various words in the language and is the foundation for applications like Information Retrieval, POS Tagging, Chunking and ultimately Machine Translation[3,4]. The available Morphological Analyzers for "Hindi" language follow stemming based, corpus based and paradigm based approaches. The corpus based approach[5] has the disadvantage that large volume of data needs to be processed. The stemming based approach[6] follows the technique of dividing the word to its corresponding stems and the suffix or prefix and then process these. To analyze a word it must have the corresponding stems in its dictionary. This approach may successfully analyze regular words but it cannot analyze irregular words. A paradigm defines all the word forms of a given stem and also provides a feature structure with every word form. The paradigm approach[7] alone cannot successfully analyze the morphology of the word-forms because of the inflection by the circumfix and gives better results when combined with one of the other approaches.

The approach for designing Morphological Analyzers is dominated by those for agglutinative languages i.e. languages like English that show low degree of inflection. Though agglutinative languages show high morpheme per word ratio and have complex morphotactic structures, the absence of fusion at morpheme boundaries makes the task of segmentation fluent once the model for implementation of morphotactics is ready. On this background, a morphological analyzer for highly inflectional language like Hindi which has the tendency to overlay the morphemes in a way that aggravates the task of segmentation presents an interesting case study[8].

## 2. Methodology

The development of morphological analyzer consists of two phases – lexicon generation and generation of morphological processor, which is done using the SFST tool. This morph-analyzer was then tested for its effectiveness by using it in a Hindi parts of speech (POS) tagger which was developed through the API provided by Stanford POS Tagger.

SFST was developed by the Institute for Natural Language Processing, University of Stuttgart. It comprises a compiler, which translates finite state transducer programs into minimized transducers and a wide range of transducer . Also, it supports UTF-8 character coding which is important for the implementation of Hindi computational morphologies[9,10].

POS Tagger for Hindi language is developed using the Stanford POS Tagger that provides a Java API. This API follows the log-linear approach for POS tagger[11,12].

### 2.1 Lexicon Generation

The lexicon was generated with the help using the raw corpus collected from the LDC-IL (Linguistic Data Consortium) for Indian Languages. Fig. 1 shows the approach used for lexicon generation. Unique words from the corpus are extracted and sorted to make the task of processing of the words manually easier. These words are then manually classified into various classes and according to their inflection, and derivations types. The total number of base words contained in the classified lexicon files is 91930. Classified lexicon file of the base words for nouns contains 50520 base words. The lexicon file of pronouns contains 81 base words. The lexicon file of adjectives contains 33006 base words. The lexicon file of verbs contains 8513 base words. The lexicon file of adverbs contains 1559 base words. The lexicon file of particles contains 12 base words. There are 9930 base words for both adjectives and nouns.

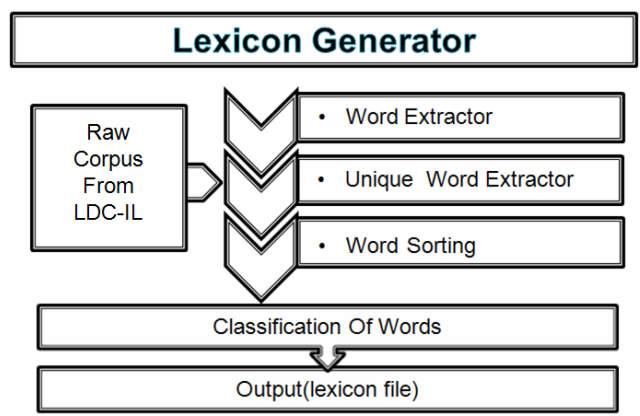

Fig. 1  Lexicon Generator.

### 2.2 Morphological Processor

Fig.2 illustrates the approach used for developing the morphological processor which has two components - the Morphological Analyzer and Morphological Generator. Both analyzer and generator require a dictionary of the root words, file containing FST rules, and the dictionary of indeclinable words. The inflectional and derivational rules are hand written and then coded with the help of SFST tool to generate the .fst files. These .fst files combined with the lexicon generate the Finite State Transducer. The dictionary indeclinable words contains the words that have some specific grammatical word structure such as अतःकरण. Some specific different rules need to be written for these indeclinable words. The analyzer takes as input the surface form and produces the result as the grammatical structure of the word or the lexicon form. The Generator takes as input the lexicon form and produce the corresponding surface form.

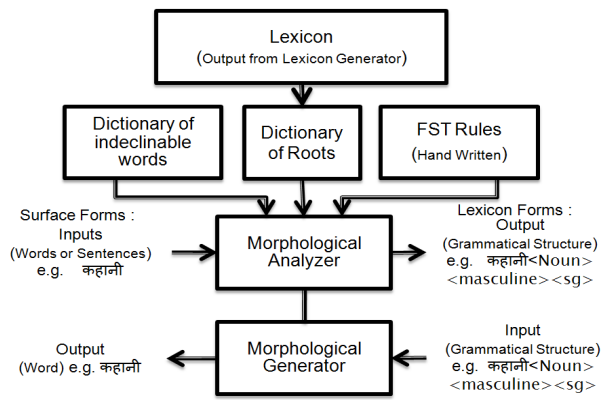

Fig. 2  Morphological Processor.

### 2.3 POS Tagger

Fig 3 illustrates the system model for the POS Tagger which contains two modules – the training module and the tagger module. The training of the model file has been done by collecting the text corpus from the Hindi news paper website Dainic Jagaran and then tagging the corpus manually. This tagged corpus is used to train the tagger resulting in the generation of the trained model file and dictionary words. The trained model file contains all the probable part-of-speech that may be assigned to each word. The dictionary contains all the root words from the corpus.

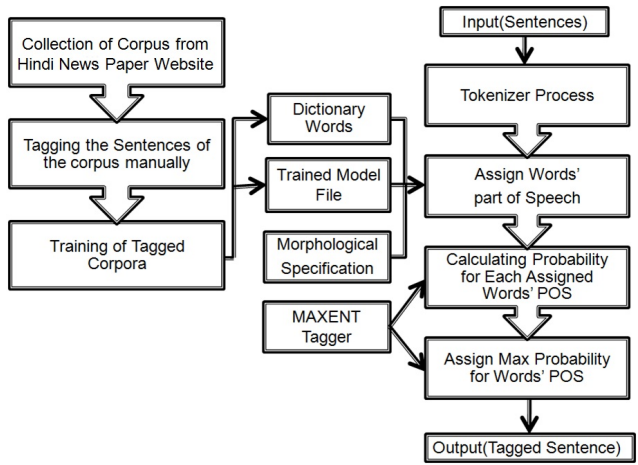

Fig. 3 POS Tagger

For tagging of a Hindi sentence, the firstly the Hindi sentence is tokenized. Each of these tokens is then searched in the dictionary and assigned all the probable parts-of-speech as per the trained model file. If the token is not a root word then the morphological specification provides the way to find out the base word-form(s) for that token. MAXENT Tagger (provided along with Stanford Tagger) is then use to calculate the probabilities of each of the parts-of-speech for a given token. The token is tagged with the part-of-speech with maximum probability.

## 3. Experimental Results

For testing the morph-analyzer and the POS Tagger an interface was developed using Java technology. This interface contains the keyboard that is provided by Google Hindi Keyboard. Which is available under Google Transliteration IME that provides an input method editor which allows users to enter text in one of the supported languages using a roman keyboard.

### 3.1 Morphological Analyzer

The morphological analyzer was tested with about 4000 inflectional, derivational and compound words. Some of the sample results are shown in tables 1, 2, 3.

Table 1 Results for Noun Inflections

| Input Words | Inflection Type | Output |
|---|---|---|
| लड़का लड़की | Gender Inflection | लड़का<Noun><masculine><sg>लड़की<Noun><feminine><sg> |

Table 2 Results for Noun Derivations

| Input Words | Inflection Type | Output |
|---|---|---|
| माली मालन | Gender Inflection | माली<Noun><masculine><sg><br>माली<Noun><feminine><sg> |
| कहानी कहानियाँ | Number Inflection | कहानी<Noun><masculine><sg><br>कहानी<Noun><masculine><pl> |
| अरे लड़के | Case Inflection | लड़का<Noun><Vocative> |
| मेज़ मेज़े | Number Inflection | मेज़<Noun><Masculine><sg><br>मेज़<Noun><Masculine><pl> |
| शेर शेरनी | Gender Inflection | शेर<Noun><Masculine><sg><br>शेर<Noun><feminine><sg> |

| Input Words | Derivation from | Output |
|---|---|---|
| शर्म बेशर्म | Noun | शर्म<Noun><Masculine><sg><br>बेशर्म<Noun><Masculine><sg> |
| मीठा मिठाई | Adjective | मीठा<Noun><Masculine><sg>मिठाई<Noun><Masculine><sg> |
| कमीना कमीनापन | Adjective | कमीना<Noun><Masculine><sg><br>कमीनापन<Noun><Masculine><sg> |
| पवित्र पवित्रता | Adjective | पवित्र<Noun><Masculine><sg><br>पवित्रता<Noun><Masculine><sg> |

Table 3 Results for Verb Inflection

| Input Words | Inflection Type | Output |
|---|---|---|
| जा रहा | Person Inflection | जा<Verb><Indicative><Masculine><Progressive><sg> |
| जा रहे | Person Inflection | जा<Verb><Indicative><Masculine><Progressive><pl> |
| पढ पढी | Gender Inflection | पढ\<Verb\>\<Indicative\>\<Masculine\>पढ\<Verb\>\<Indicative\>\<Faminine\> |
| जा जातें | Number Inflection | जा\<Verb\>\<present\>जा\<Verb\>\<Transitive\>जा\<Verb\>\<Dative\>जा\<Verb\>\<Imprative\>\<Intimate\>जा\<Verb\>\<Indicative\>\<Masculine\>\<Perfectiv\>\<sg\> |
| करता करते | Number Inflection | कर\<Verb\>\<Indicative\>\<Masculine\>\<Habitual\>\<sg\>कर\<Verb\>\<Indicative\>\<Masculine\>\<Habitual\>\<pl\> |

Table 4 depicts the performance of the Morph-Analyzer.

Table 4 Performance Evaluation for Morph Analyzer

| Word Input Type | No. of Words | Percentage of Correct Results |
|---|---|---|
| Noun Inflections,Derivations, Compounds | 1000 | 93 |
| Adjective Inflections,Derivations, Compounds | 1000 | 95 |
| Verb Inflections,Derivations, Compounds | 1000 | 92 |
| Adverb Inflections,Derivations, Compounds | 1000 | 98 |

### 3.2 POS Tagger

The Part-of-Speech (POS) Tagger was tested with about 100 sentences. Some of the sample results are shown in tables 4, 5.

Table 5 Results for Simple Sentence Tagging

| Sentence Input | Tagged Sentence |
|---|---|
| मैं घर जा रहा हूँ। | मैं/PR_PRI घर/N_NN जा/V_VM रहा/V_AUX हूँ/V_AUX ।/। |
| यह सभी अलंकरण पंक्तिबढ़ होकर संपूर्ण मंदिर के एक माला के सदृश घेरे हैं। | यह/PR_PRI सभी/JJ अलंकरण/N_NN पंक्तिबढ़/JJ होकर/V_VM संपूर्ण/JJ मंदिर/N_NN के/PSP एक/QT_QTC माला/N_NN के/PSP सदृश/JJ घेरे/N_NN हैं/V_AUX ।/। |

Table 6 Results for the POS Tagger of Tagging the Sentence with Ambiguity words

| Sentence Input | Tagged Sentence |
|---|---|
| आम आदमी आम खाता है। | आम/JJ आदमी/N_NN आम/N_NN खाता/V_VM है/V_AUX ।/। |
| उसका खाता संख्या एक है। | उसका/PR_PRI खाता/N_NN संख्या/JJ एक/QT_QTC है/V_AUX ।/। |
| आम आदमी आम बेचता है। | आम/JJ आदमी/N_NN आम/N_NN बेचता/V_VM है/V_AUX ।/। |

Table 7 depicts the performance of POS Tagger for Hindi Language.

Table 7 Performance table of POS Tagger

| Sentence Input | No. of Sentences | Accuracy of Correct Results (In Percentage) |
|---|---|---|
| Sentence with Known Words | 100 | 93 |
| Sentence with Unknown Words | 100 | 87 |

### 4. Conclusion

The Morph-Analyzer was successfully developed and used in POS tagger. It can, similarly, be used in other NLP applications. The Morph Analyzer can be enhanced by combining the paradigm approach with the FST approach.